\documentclass{article}

    \usepackage[preprint]{neurips_2025}

\usepackage[utf8]{inputenc} % allow utf-8 input
\usepackage[T1]{fontenc}    % use 8-bit T1 fonts
\usepackage{hyperref}       % hyperlinks
\usepackage{url}            % simple URL typesetting
\usepackage{booktabs}       % professional-quality tables
\usepackage{amsfonts}       % blackboard math symbols
\usepackage{nicefrac}       % compact symbols for 1/2, etc.
\usepackage{microtype}      % microtypography
\usepackage{xcolor}         % colors

\usepackage{tcolorbox}
\usepackage{pifont}
\usepackage{booktabs}
\usepackage{multicol}
\usepackage{multirow}
\usepackage{adjustbox}
\usepackage{makecell}
\usepackage{bbding}
\usepackage{etoolbox}
\usepackage{tikz}
\usepackage[para,online,flushleft]{threeparttable}
\usepackage{xspace}
\usepackage{duckuments}
\usepackage{graphicx}
\usepackage{url}
\usepackage{mdframed}
\usepackage{multicol}
\usepackage{multirow}
\usepackage{colortbl}
\usepackage{tabularx}
\usepackage{wrapfig}
\usepackage{subcaption}
\usepackage{tabularx}
\usepackage{caption}
\usepackage{float}
\usepackage{amsmath}
\usepackage[capitalize,noabbrev]{cleveref}
\usepackage{mathtools}

\usepackage{enumitem}
\setlist[itemize]{leftmargin=*}

\newcommand{\methodname}{\texttt{UProp}\xspace}
\DeclarePairedDelimiterX{\infdivx}[2]{(}{)}{%
  #1\;\delimsize\|\;#2%
}
\newcommand{\infdiv}{D_{\textit{KL}}\infdivx}

\usepackage{amsmath,amsfonts,bm}

\def\eqref#1{equation~\ref{#1}}
\def\1{\bm{1}}

\def\vtheta{{\bm{\theta}}}

\def\vc{{\bm{c}}}

\def\vo{{\bm{o}}}

\def\vx{{\bm{x}}}
\def\vy{{\bm{y}}}

\DeclareMathAlphabet{\mathsfit}{\encodingdefault}{\sfdefault}{m}{sl}
\SetMathAlphabet{\mathsfit}{bold}{\encodingdefault}{\sfdefault}{bx}{n}

\newcommand{\Var}{\mathrm{Var}}

\title{\methodname: Investigating the Uncertainty Propagation of LLMs in Multi-Step Agentic Decision-Making}

\author{%
 \bf Jinhao Duan$^1$\,\, James Diffenderfer$^{2}$\,\, Sandeep Madireddy$^{3}$\,\, \\ \bf Tianlong Chen$^{4}$ \,\, Bhavya Kailkhura$^{2}$ \,\, Kaidi Xu$^1$\thanks{Corresponding to: Kaidi Xu \texttt{kx46@drexel.edu}}
   \\
  $^1$Drexel University 
  $^2$Lawrence Livermore National Laboratory \\
  $^3$Argonne National Laboratory
  $^4$UNC Chapel Hill 
}

\newtheorem{theorem}{Theorem}

\begin{document}

\maketitle

\begin{abstract}

As Large Language Models (LLMs) are integrated into safety-critical applications involving sequential decision-making in the real world, it is essential to know when to trust LLM decisions. Existing LLM Uncertainty Quantification (UQ) methods are primarily designed for single-turn question-answering formats, resulting in multi-step decision-making scenarios, e.g., LLM agentic system, being underexplored.
In this paper, we introduce a principled, information‑theoretic framework that decomposes LLM sequential decision uncertainty into two parts: (i) internal uncertainty intrinsic to the current decision, which is focused on existing UQ methods, and (ii) extrinsic uncertainty, a Mutual-Information (MI) quantity describing how much uncertainty should be inherited from preceding decisions. We then propose \methodname, an efficient and effective extrinsic uncertainty estimator that converts the direct estimation of MI to the estimation of Pointwise Mutual Information (PMI) over multiple Trajectory-Dependent Decision Processes (TDPs).
\methodname is evaluated over extensive multi-step decision-making benchmarks, e.g., AgentBench and HotpotQA, with state-of-the-art LLMs, e.g., GPT-4.1 and DeepSeek-V3. Experimental results demonstrate that \methodname significantly outperforms existing single-turn UQ baselines equipped with thoughtful aggregation strategies. Moreover, we provide a comprehensive analysis of \methodname, including sampling efficiency, potential applications, and intermediate uncertainty propagation, to demonstrate its effectiveness. Codes will be available at~\url{https://github.com/jinhaoduan/UProp}.
\end{abstract}

\section{Introduction}

% background: hallucination, know when to trust, llm multi-step decision making
Large Language Models (LLMs)~\cite{zhao2023survey} are increasingly deployed in real-world applications that involve sequential decision-making, such as Agentic AI~\cite{wang2024survey}, where LLMs interact with environments across multiple steps. Many of these applications, including multi-round medical consultations~\cite{zhou2023survey} and autonomous robotic control~\cite{zeng2023large,duan2022survey}, are safety-critical. Given that LLMs are prone to hallucinations and errors~\cite{huang2025survey}, it is crucial to assess the reliability of their decisions and understand when these decisions can be trusted. Uncertainty quantification (UQ) estimates the degree of uncertainty or lack of confidence that a model has in its predictions, essentially reflecting how unsure it is about the ``correctness'' of its output~\cite{gawlikowski2023survey}. It has proven to be a promising method for quantifying the reliability of LLM decisions, such as in hallucination detection and correction~\cite{yin2024reasoning}.

% Existing method: baselines, single-turn, intrinsic, multi-step more challenging and complex
Current LLM UQ methods primarily focus on single-step question-answering tasks~\cite{malinin2020uncertainty}, where LLMs are expected to respond to a query. These methods~\cite {kuhnsemantic,duan2024shifting,lingenerating,qiu2024semantic} quantify uncertainty by measuring the semantic diversity of LLM output space. While these ``single-step'' methods offer reliable uncertainty estimations at each step, in the multi-step decision-making scenarios, they fail to capture the propagation of uncertainty within a decision trajectory. SAUP~\cite{zhao2024saup} trains a Hidden Markov Model (HMM) to predict the aggregation weights of per-step uncertainty within a decision trajectory. However, it requires the ground-truth labels from the test domain and does not investigate uncertainty propagation in a principled manner. In this paper, we study
{\textbf{\textit{how the uncertainty of the current decision should be influenced by preceding decisions?}}}

\begin{figure}[t]
    \centering
    \includegraphics[width=\linewidth]{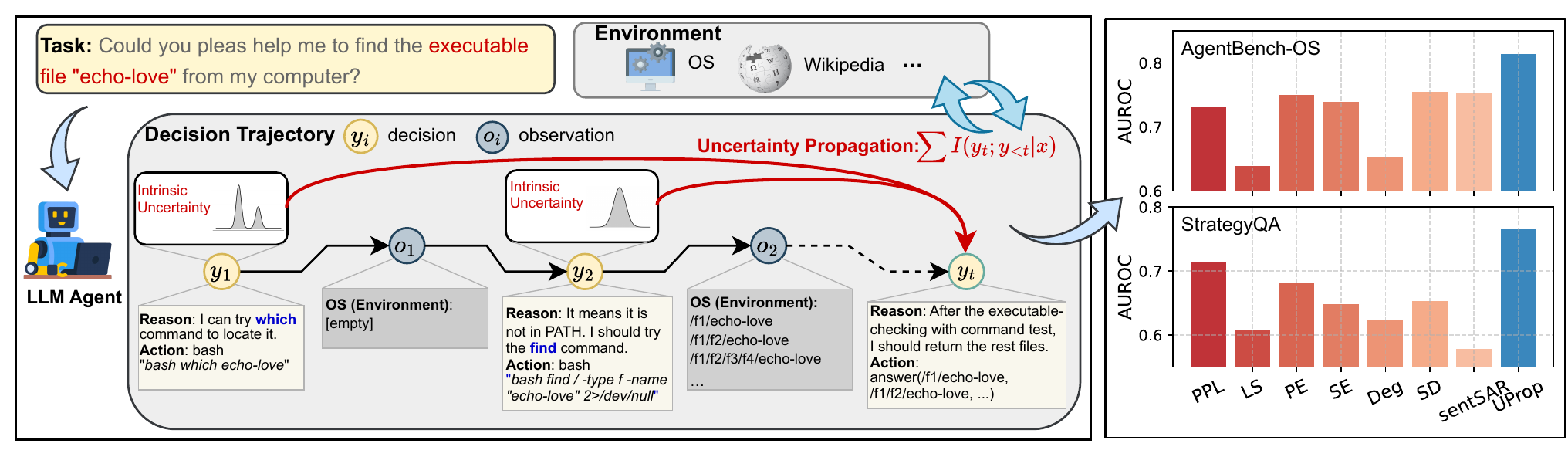}
    \caption{Overall pipeline of uncertainty propagation of LLMs 
 in multi-step agentic decision-making. }
    \label{fig:uncertainty_prop}
    \vspace{-5mm}
\end{figure}

We approach LLM multi-step decision-making from a Bayesian perspective and develop an information-theoretic framework to analyze its uncertainty propagation. Specifically, we decompose the LLM's uncertainty at each decision step into (i) \textbf{Intrinsic Uncertainty} (IU), which reflects the \underline{internal uncertainty} dependent solely on the current state, e.g., observation the LLM is facing, and (ii) \textbf{Extrinsic Uncertainty} (EU), which represents the uncertainty introduced by (or ``inherited from'') the \underline{variability of preceding decisions}. Among the two components, intrinsic uncertainty can be reliably estimated, as it is convenient to sample from the decision distribution from LLMs for uncertainty estimation~\cite{malinin2020uncertainty}. In contrast, estimating extrinsic uncertainty is more challenging because it is described as the Mutual Information (MI) between the current decision distribution and each of the preceding decision distributions~\cite{cover1999elements}. This becomes intractable as MI necessitates the decision distributions at each step of the reasoning process. Even from a Monte Carlo (MC) sampling perspective, the multi-step decision-making scenario demands the exploration of an \underline{exponentially expanding decision space}~\cite{kraskov2004estimating}, which is computationally infeasible.

We propose \methodname, an efficient and effective estimator of extrinsic uncertainty. 
In general, \methodname complies with the MC approximation idea which first samples decision processes from the decision space and then estimates per-process uncertainty propagation. 
Specifically, \methodname \ding{202} first conducts Trajectory-Dependent Decision Process (TDP) sampling from the exponential decision space: each TDP sample results in a complete decision trajectory (from beginning to end) along with multiple samples at each step.
\ding{203} Then, for each TDP, \methodname estimates the uncertainty propagation by approximating the more feasible Pointwise MI (PMI).
With convergence analysis, we prove that the Trajectory-Dependent PMI approximation converges to the actual MI in the LLM multi-step decision-making scenario, under a mild local smoothness assumption.

\methodname is evaluated in extensive LLM multi-step decision-making scenarios including the Operating System Agent split in AgentBench~\cite{liu2023agentbench} and multi-hop benchmarks such as HotpotQA~\cite{yang2018hotpotqa} and StrategyQA~\cite{geva2021did} bound with Wikipedia engine, over powerful LLMs, such as GPT-4.1-Nano~\cite{achiam2023gpt}, GPT-3.5-Turbo~\cite{brown2020language}, DeepSeek-V3~\cite{DeepSeekAI2024DeepSeekV3TR}. We compare \methodname with state-of-the-art single-turn UQ methods, including Semantic Entropy (SE)~\cite{kuhnsemantic}, Deg~\cite{lingenerating}, SAR~\cite{duan2024shifting}, Semantic Density (SD)~\cite{qiu2024semantic}, etc., equipped with thoughtful step aggregation strategies. Experimental results demonstrate that \methodname significantly outperforms these baselines (by 2.3\% $\sim$ 11\% AUROC). We further characterize \methodname from the perspective of sampling efficiency, selective prediction, and intermediate uncertainty propagation. Our results indicate that extrinsic uncertainty plays an important role in the uncertainty quantification of LLM sequential decision-making. 
Our contributions can be summarized as follows:
\begin{itemize}
    \item We provide an information-theoretic framework that decomposes the uncertainty of LLM sequential decision into intrinsic and extrinsic uncertainty. We highlight the necessity of propagating extrinsic uncertainty along the LLM decision chain for more accurate uncertainty quantification.
    \item We provide \methodname, an efficient and effective extrinsic uncertainty estimator. \methodname approximates the Mutual Information (MI) between decision distributions by expectating the Pointwise Mutual Information (PMI) among trajectory-dependent samplings.
    \item \methodname is evaluated over extensive LLM sequential decision-making scenarios, involving powerful LLMs and state-of-the-art baseline methods. Experimental results demonstrate that \methodname significantly outperforms best-performing baselines in LLM multi-step decision-making scenarios.
\end{itemize}

\section{Preliminary}

\subsection{Uncertainty Quantification in Auto-Regressive Generations}
From the Bayesian perspective, UQ measures the uncertainty within the predictive probability distribution $p_{\vtheta}(\vy|\vx)$ over the LLM output space $\mathcal{Y}$, given a parameterized LLM $f_{\vtheta}$ and instruction $\vx$.
One of the most popular UQ methods is quantifying the total uncertainty of the predictive distribution by calculating its Predictive Entropy (PE)~\cite{gawlikowski2023survey}. However, considering that the analytic form of LLM predictive distributions is intractable, i.e., do not have access to all possible $|V|^k$ $k$-length generations in the LLM output space (where $V$ is the vocabulary size), a more convenient way is approximating via Monte-Carlo (MC) sampling~\cite{malinin2020uncertainty}:
$$
PE(\vx) = H(\vy|\vx) = \int{p_{\vtheta}(\vy|\vx)\log(p_{\vtheta}(\vy|\vx))} \, d\vy \approx - \frac{1}{N}\sum_{i}^{N} {\log{p_{\vtheta}(\vy^{(i)}|\vx)}}, \,\, \vy^{(i)} \sim p_{\vtheta}(\vy|\vx),
$$

where $N$ is the number of samples and $p_{\vtheta}(\vy^{(i)}|\vx)=\prod_{i}^{L_i} p_{\vtheta}(z_i|z_{<i}, \vx)$ is the generative probability of $\vy^{(i)}$ with length $L_i$. $z_i$ is the $i$-th token of $\vy^{(i)}$. Length-normalization is also commonly applied to mitigate the length sensitivity: $LN\text{-}PE(\vx) \approx - \frac{1}{N}\sum_{i}^{N} {\frac{1}{L_i}\log{p_{\vtheta}(\vy^{(i)}|\vx)}}$.
Furthermore, \citet{kuhnsemantic} proposes that PE may overestimate output uncertainty due to the existence of semantic clusters, i.e., different generations may share the same semantics. To mitigate this, Semantic Entropy (SE) calculates the cluster-wise predictive entropy with MC approximation:
$$
SE(\vx) \approx -\frac{1}{C}\sum_{i}^{C}{\log(p_{\vtheta}(\vc_i|\vx))}, \,\, p_{\vtheta}(\vc_i|\vx)=\sum_{\vy \in \vc_i}{p_{\vtheta}(\vy|\vx)},
$$
where $C$ is the number of semantic clusters and $\vc_i$ is the $i$-th cluster consisting of generations $\vy_i$ sharing the same semantics. Following the semantic consistency, a series of UQ methods including Deg~\cite{lingenerating}, SAR~\cite{duan2024shifting}, and SD~\cite{qiu2024semantic} are proposed.

\subsection{LLM Multi-Step Agentic Decision-Making}
LLM multi-step agentic decision-making~\cite{liu2023agentbench,duan2024gtbench} is usually modeled as a stochastic Markov Decision Process (MDP)
$(f_\vtheta, \mathcal{O}, \mathcal{Y}, \mathcal{T})$, where LLM $f_\vtheta$ interacts with the environment continuously. $\mathcal{O}$ and $\mathcal{Y}$ are observation space and decision space, respectively. $\mathcal{T}: \mathcal{Y^*} \rightarrow \mathcal{O}$ is the deterministic observation transition function of the environment, where $\mathcal{Y}^*$ denotes a finite sequence of decisions. Assume at the $t$-th decision step, the decision $\vy_t \in \mathcal{Y} $ is sampled by
$$
\vy_t \sim p_{\vtheta}(\vy_t|\vo_{t-1}, \vy_{t-1}, \cdots, \vo_1, \vy_1, \vx),
$$
where $\vo_i \in \mathcal{O}$ is the observation at $i$-th step and $\vx$ is the instruction. We assume the observation transition function is deterministic when the preceding decisions $\mathcal{Y}^*=[\vy_i]^{t-1}$ are determined, i.e., the decision distribution $\vy_t$ is solely dependent on preceding decisions. Thus, we omit all the observation conditions in the following notations, i.e., $\vy_t \sim p_{\vtheta}(\vy_t|\vy_{1:t-1}, \vx)$.

\section{Methodology}
\subsection{Predictive Uncertainty Propagation in LLM Multi-Step Decision-Making}
In the LLM multi-step decision-making process, UQ quantifies the uncertainty within the predictive distribution $p_{\vtheta}(\vy|\vx)$.
Without loss of generality, we quantify the uncertainty at the $t$-th step predictive distribution $\vy_t \sim p_{\vtheta}(\vy_t|\vx)$. By marginalizing preceding decisions, we obtain the following decomposition (see~\cref{appendix:uncertainty_decompose} for detailed procedures):

\begin{equation}\label{eq:factorization}
\begin{aligned}
p_{\vtheta}(\vy_t|\vx) & = \int{p_{\vtheta}(\vy_{t}|\vy_{t-1}, \vx)p(\vy_{t-1}|\vx)} d\vy_{t-1} \\
    & = \int{\underbrace{p_{\vtheta}{(\vy_t|\vy_{1:t-1}, \vx})}_{\text{\small Intrinsic Uncertainty}}}\underbrace{\prod\nolimits_{i}^{t-1}{p_{\vtheta}(\vy_i|\vy_{1:i-1}, \vx)}}_{\text{\small Extrinsic Uncertainty}}d\vy_1 d\vy_2 \cdots d\vy_{i-1}. 
\end{aligned}
\end{equation}
We show that the total uncertainty at step $t$ could be described in \textbf{Intrinsic Uncertainty} (IU) and \textbf{Extrinsic Uncertainty} (EU): (1) IU refers to the expected variance of $\vy_t$ given all preceding decisions, i.e., $\mathbb{E}_{\vy_{1:t-1}}{\big[\Var_{\vy_t}(\vy_t|\vy_{1:{t-1}}, \vx)\big ]}$. It captures the uncertainty within the predictive distribution itself and corresponds to what ``single-step'' UQ methods typically estimate; (2) EU quantifies the variance of $\vy_t$ introduced by prior decisions, expressed as $\Var_{\vy_{1:t-1}}\big(\mathbb{E}_{\vy_t}[\vy_t|\vy_{1:t-1}, \vx]\big )$, which is the uncertainty that should be propagated from preceding decisions.

By the chain rule of conditional entropy, entropy $H(\vy_t|\vx)$ could be expressed as the following (see \cref{appendix:entropy_decom} for detailed procedures):
\begin{equation}\label{eq:entropy_factorization}
    \begin{aligned}
        H(\vy_t|\vx) & = H(\vy_t|\vy_{1:t-1}, \vx) + \sum\nolimits_{i}^{t-1} {(H(\vy_{t}|\vx) - H(\vy_{t}|\vy_{i}, \vx))} \\
        & = \underbrace{H(\vy_t|\vy_{1:t-1}, \vx)}_{\text{\small  Intrinsic Uncertainty}} + \underbrace{\sum\nolimits_{i}^{t-1} {I(\vy_t;\vy_i|\vy_{i+1:t-1}, \vx)}}_{\text{\small  Extrinsic Uncertainty}},
    \end{aligned}
\end{equation}
where $
I(\vy_t;\vy_i|\vy_{i+1:t-1}, \vx)$ is Mutual Information (MI). The total uncertainty of the decision process $\mathcal{P}=(\vy_1, \vy_2, \cdots, \vy_t) \sim p_{\vtheta}{(\mathcal{P}|\vx)}$ becomes: 
\begin{equation}\label{eq:P_total_uncertainty}
    H(\mathcal{P}) = \mathbb{E}_{P \sim \mathcal{P}} \Big[ -\log {p_{\vtheta}(P|\vx)} \Big] = \mathbb{E}_{P \sim \mathcal{P}} \Big[ - \sum_{i} {\log p_{\vtheta}{(\vy_i|\vy_{1:i-1}, \vx)}} \Big],
\end{equation}
Within the decomposition in~\cref{eq:entropy_factorization},
\begin{itemize}
    \item Intrinsic Uncertainty could be conveniently MC approximated by first sampling multiple generations from $p_{\vtheta}(\vy_t|\vy_{1:t-1}, \vx)$ and then aggregating with existing algorithms, such as PE, SE, and SAR.
    \item Extrinsic Uncertainty is characterized by the cumulative MI between $\vy_t$ and all preceding decisions $\vy_{1:t-1}$. It reflects the extent to which uncertainty in $\vy_t$ is reduced as each prior decision is resolved, i.e., \textit{knowledge uncertainty}~\cite{malinin2019uncertainty}. In this sense, extrinsic uncertainty quantifies the degree of ``increased determinism'' in $\vy_t$ that arises from conditioning on $\vy_{1:t-1}$.
\end{itemize}

However, directly calculating extrinsic uncertainty is intractable, as it requires an awareness of predictive distributions.
Even from the perspective of MC approximation or density estimation, the estimation of $I(\vy_t;\vy_i|\vy_{i+1:t-1}, \vx)$ is still challenging as it necessarily explores an exponentially spanned decision space: outer sampling from preceding decision distributions $\vy_{<t}$ with inner sampling from $\vy_{t}$. Moreover, in the LLM decision-making scenarios, this exponential interaction with the environment becomes harder to afford. Please refer to~\cref{appendix:exponential_explore} for more discussion.

\subsection{\methodname: Estimate Extrinsic Uncertainty with Trajectory-Dependent Pointwise MI}

We propose \methodname as an efficient and effective estimator of EU. 
In general, \methodname complies with the MC approximation idea which first samples decision processes from the decision
space and then estimates per-process uncertainty propagation. Specifically, it converts the direct estimation of MI to the estimation of \textit{Pointwise Mutual Information} over \textit{Trajectory-Dependent Decision Processes}: 

\textbf{Trajectory-Dependent Decision Process (TDP) Sampling} \,\,
Starting from the beginning decision step ($t=1$), we first sample $N$ decisions $\{\vy_{t}^{(1)}, \vy_{t}^{(2)}, \cdots, \vy_{t}^{(N)}\} \sim p_{\vtheta}(\vy_t|\vx)$; then, we randomly select one sample $\vy_{t}^{(k)}$ by probability, as the preceding realization of the $(t+1)$-th step; then, we sample $N$ decisions from $\vy_{t+1} \sim p_{\vtheta}{(\vy_{t+1}|\vy_{1:t}=\vy_{1:t}^{(k)}, \vx)}$. We repeat this protocol until $\vy_{T}^{(k)}$ achieves an end decision at step $T$, e.g., the decision to return the final answer. In this way, each TDP will be expressed as:
$$
\textit{TDP}_{z} = \{<\vy_{1}^{(k)}, \{\vy_{1}^{(n)}\}_{n, n\neq k}^N>, <\vy_{2}^{(k)}, \{\vy_{2}^{(n)}\}_{n, n\neq k}^N>, \cdots, 
<\vy_{T}^{(k)}, \{\vy_{T}^{(n)}\}_{n, n\neq k}^N> \},
$$
consisting of one complete decision trajectory: $\{\vy_{1}^{(k)}, \vy_{2}^{(k)}, \cdots, \vy_{T}^{(k)}\}$, and multiple samplings conditioned on preceding trajectories at each step: $\{ \{\vy_{1}^{(n)}\}_{n, n\neq k}^N, \{\vy_{2}^{(n)}\}_{n, n\neq k}^N, \cdots, \{\vy_{T}^{(n)}\}_{n, n\neq k}^N\}$.

\textbf{Pointwise Mutual Information (PMI) in TDP} \,\, We study the uncertainty propagation over TDPs. Conditioned on the realizations within TDP, the MI $I(\vy_{t};\vy_{t-1}|\vx)$ over TDP at step $t$ becomes a PMI:
\begin{equation}\label{eq:PMI}
    \textit{PMI}(\vy_{t};\vy_{t-1}=\vy_{t-1}^{(k)}|\vx) = \infdiv{p_{\vtheta}(\vy_t|\vy_{t-1}^{(k)}, \vx)}{p_{\vtheta}(\vy_t|\vx)}.
\end{equation}
Then, the MC approximated total uncertainty of TDP, $\mathcal{P}_{\textit{TDP}} \sim p_{\vtheta}(\mathcal{P}_{\textit{TDP}}|\vx)$, becomes
\begin{equation}\label{eq:TDP_total_uncertainty}
    H(\mathcal{P}_{\textit{TDP}}) \approx \frac{1}{Z} \sum_z^Z \sum_{t}^{T_z} {\Big( H(\vy_t|\vy_{1:t-1}^{(k)}, \vx) + \sum_{i}^{t-1} \textit{PMI}(\vy_t;\vy_i^{(k)}|\vy_{i+1:t-1}^{(k)}, \vx)\Big)},
\end{equation}
where $Z$ is the sampling number of TDP and $T_z$ is the number of steps within the $z$-th TDP, i.e., the length of the $z$-th TDP's decision trajectory.
\begin{theorem}\label{theorem:MI_converge} (Convergence of the TDP Sampling) With sufficiently large TDP sampling, the total uncertainty of TDP converges to the total uncertainty $H(\mathcal{P})$~(\cref{eq:P_total_uncertainty}): 
$H(\mathcal{P}_{\textit{TDP}}) \rightarrow H(\mathcal{P})$, when $Z \rightarrow \infty \ $.
\end{theorem}
Please refer to~\cref{appendix:proof_MI_converge} for the proof of~\cref{theorem:MI_converge}.
Instead of directly estimating MI over the exponential decision space, \methodname first samples \underline{linear-spanning} decision processes and then uses the more feasible PMI over each TDP as the approximation of the total uncertainty $H(\mathcal{P})$.

\subsection{Spreading Decision Distributions by Preceding 
Variance}\label{sec:PMI_estimation}
Given a TDP $P_{z}$, we consider approximating PMI by spreading from the known conditional distribution, under a mild local smoothness assumption. Specifically, without loss of generality, we consider the MC approximation of $\textit{PMI}(\vy_{t};\vy_{t-1}^{(k)}|\vx)$:
\begin{equation}
    \textit{PMI}(\vy_{t};\vy_{t-1}=\vy_{t-1}^{(k)}|\vx) = \mathbb{E}_{\vy_t} \Big[ \log \frac{p_{\vtheta}(\vy_{t}|\vy_{t-1}^{(k)}, \vx)}{p_{\vtheta}(\vy_t|\vx)}\Big] \approx \frac{1}{N} \sum_{n}^{N} \log \frac{p_{\vtheta}(\vy_{t}^{(n)}|\vy_{t-1}^{(k)}, \vx)}{p_{\vtheta}(\vy_{t}^{(n)}|\vx)},
\end{equation}
where $\vy_{t}^{(n)}$ is the $n$-th sample from TDP's $t$-step samples and $p_{\vtheta}(\vy_{t}^{(n)}|\vy_{t-1}^{(k)}, \vx)$ is calculated and saved during TDP sampling. In terms of $p_{\vtheta}(\vy_{t}^{(n)}|\vx)$, we approximate it by spreading the preceding semantic variance with ``neighborhood-weighted'' average:
\begin{equation}
    \hat{p}_{\vtheta}(\vy_t|\vx) = \sum_{n}^N p_{\vtheta}(\vy_{t}|\vy_{t-1}^{(k)}, \vx) \cdot K_{N}(d(\vy_{t-1}^{(n)}, \vy_{t-1}^{(k)})),
\end{equation}
where $K_{\tau}(x)=\big(\frac{1}{\sqrt{2\pi}}e^{(-\frac{x^2}{2})}\big)^\tau$ is a Gaussian Kernel with $\tau$ controls its sharpness. $d(\vy_1, \vy_2)$ is a distance measurement between the two decisions. We take $s=N$ to highlight those samples close to $\vy_{t-1}^{(k)}$, i.e., the extrinsic uncertainty is dominated by its surroundings.
In this way, the PMI is approximated as:
\begin{equation}\label{eq:PMI_approx}
    \widehat{\textit{PMI}}(\vy_{t};\vy_{t-1}^{(k)}| \vx) = \frac{1}{N} \sum_{n}^{N} \log \frac{p_{\vtheta}(\vy_{t}^{(n)}|\vy_{t-1}^{(k)}, \vx)}{\hat{p}_{\vtheta}(\vy_{t}^{(n)}|\vx)} = -\log{\sum_{n}^{N} K_N(d(\vy_{t-1}^{(n)}, \vy_{t-1}^{(k)}))}.
\end{equation}
Heuristically, spreading by preceding semantic variance indicates that a low-uncertainty preceding decision distribution introduces less uncertainty to the current step. Extremely, a degenerate preceding distribution introduces no uncertainty to follow-up decisions.

\begin{theorem}\label{theorem:PMI_convergence} (Convergence of the PMI Approximation)
    Assume that $p_{\vtheta}(\vy_{t}|\vy_{t-1}, \vx)$ satisfies a local smoothness with respect to $\vy_{t-1}$, i.e., for any fixed context $\vx$, there exists a sufficiently small neighborhood around $\vy_{t-1}$ such that for all points $\vy_{t-1}'$ within this neighborhood:
    \begin{align*}
        \forall \epsilon > 0, \exists \beta > 0: |\vy_{t-1} - \vy_{t-1}'| < \beta, \,\,\, \text{then} \,\, |p_{\vtheta}(\vy_t|\vy_{t-1}, \vx) - p_{\vtheta}(\vy_t|\vy_{t-1}', \vx)| < \epsilon.
    \end{align*}
    Then, the PMI estimation (\cref{eq:PMI_approx}) spreading from the preceding variance converges to the actual MC approximation of PMI (\cref{eq:PMI}):
    $\widehat{\textit{PMI}}(\vy_t|\vy_{t-1}, \vx) \rightarrow \textit{PMI}(\vy_t|\vy_{t-1}, \vx)$.
\end{theorem}
Please refer to~\cref{appendix:proof_PMI_estimation} for the proof of~\cref{theorem:PMI_convergence} and further discussion.
The local smoothness assumption is natural and practical and has been widely conducted in existing LLM analysis~\cite{malinovskii2024pv}. Combining the total uncertainty convergence~(\cref{theorem:MI_converge}) and PMI convergence~(\cref{theorem:PMI_convergence}), the total uncertainty~(\cref{eq:P_total_uncertainty}) approximation is derived (combining~\cref{eq:TDP_total_uncertainty,eq:PMI_approx}) as:
\begin{equation}\label{eq:uprop_approx}
    H(\mathcal{P}) = H(\mathcal{P}|\vx) \approx H(\mathcal{P}_{\textit{TDP}}|\vx) =  \frac{1}{Z} \sum_z^Z \frac{1}{\lambda_z} 
 \sum_{t}^{T_z} {\Big( H(\vy_t|\vy_{1:t-1}^{(k)}, \vx) + \sum_{i}^{t-1} \widehat{\textit{PMI}}(\vy_t;\vy_i^{(k)}|\vy_{i+1:t-1}^{(k)}, \vx)\Big)},
\end{equation}
where $\frac{1}{\lambda_z}$ is an additional ``step length-normalization'' item:

\textbf{Step Length-Normalization}\,\, Similar to ``length-normalization''~\cite{malinin2020uncertainty}, due to the accumulation over $\widehat{\textit{PMI}}(\vy_t;\vy_i^{(k)}|\vy_{i+1:t-1}^{(k)}, \vx)$, \cref{eq:uprop_approx} implies the Step Length Bias: \textit{longer decision steps encodes higher extrinsic uncertainty}. The total uncertainty of a TDP is normalized by $\lambda_z = \sum_t^{T_z} \sigma_t = \sum_t^{T_z} (1 + \frac{\textit{EU}}{\textit{IU}}) = T_z + \sum_{t}^{T_z} \frac{\sum_{i}^{t-1} \widehat{\textit{PMI}}(\vy_t;\vy_i^{(k)}|\vy_{i+1:t-1}^{(k)}, \vx)}{H(\vy_t|\vy_{1:t-1}^{(k)}, \vx)}$, where $\sigma_t$ indicates the relative inflation of the uncertainty at step $t$ due to extrinsic contributions.
In this way, the step bias is mitigated, and different TDPs with varying lengths become comparable.

\cref{eq:uprop_approx} estimates the overall uncertainty of decision distributions $\mathcal{P}$. However, in some scenarios, e.g., hallucination detection, one may care more about the uncertainty of a specific prediction $\vy^*$, i.e., the uncertainty of the maximum probability class. Given model output $\vy^*$, e.g., the greedy generation, its uncertainty could be approximated as: 
\begin{equation}\label{eq:uprop_max_prob}
H(\vy^*|\vx) = H(\mathcal{P}_{\vy^*}|\vx) \approx H(\mathcal{P}_{\textit{TDP}, \vy^*}|\vx),
\end{equation}
where $\mathcal{P}_{\vy^*}$ is a decision process distribution consisting of decision processes ending with decision $\vy^*$, and $\mathcal{P}_{\textit{TDP}, \vy^*}$ is a TDP distribution consisting TDPs ending with decision $\vy^*$.

In our implementation, we calculate PE as the estimation of intrinsic uncertainty. In the rest of this paper, we denote \methodname to be $H(\mathcal{P}_{\textit{TDP}, \vy^*}|\vx)$ by default. For decision distance measurement $d$, we use the simple string fuzzy matching from thefuzz~\cite{thefuzz} as the distance measurement. Please refer to~\cref{apendix:LLM_decision_dist} for more discussion of decision distance measurement.

\section{Experiments}
In this section, we provide the detailed experimental setup and discuss the performance of \methodname in multiple settings, such as sampling efficiency and potential applications.
\subsection{Experimental Setup}
\textbf{Environments and Benchmarks} \,\, We evaluate \methodname over both multi-step decision-making and multi-step reasoning benchmarks: 
\begin{itemize}
    \item \textbf{Multi-Step Decision-Making}: we consider AgentBench-OS, the \textit{Operating System (OS)} Agent benchmark in \textit{AgentBench}~\cite{liu2023agentbench}. In AgentBench-OS, the LLM Agent is instructed to finish a task by interacting with a Linux OS, e.g., \textit{find an executable file named echo-love} (see~\cref{appendix:agentbench_os} for demonstrations and prompt templates).
    \item \textbf{Multi-Step Reasoning}: we consider the popular multi-hop question-answering benchmarks: \textit{HotpotQA}~\cite{yang2018hotpotqa} and \textit{StrategyQA}~\cite{geva2021did}. In these benchmarks, LLMs are tasked to answer a question requiring multi-hop reasoning. LLM is prompted in a ReAct~\cite{yao2022react} style: \textit{Reasoning-Action-Observation}, where each action will provide a keyword to a Wikipedia engine for retrieval (see~\cref{appendix:multihop_react} for demonstrations and prompt templates).
\end{itemize}

\begin{table*}[t]
    \centering
    \caption{AUROC results over AgentBench-Operating System and StrategyQA benchmarks. For single-turn baseline UQ methods, uncertainties are aggregated by \textbf{\textit{averaging}} over all steps.}
    \label{tab:average_baseline}
    \adjustbox{width=1\textwidth}{
    \begin{tabular}{lcccccccc|c}
\toprule
\multicolumn{1}{c}{\textbf{Models}} & \textbf{Success Rate} & PPL & LS & PE & SE & Deg & SD & sentSAR & \methodname (ours) \\
\midrule
& & \multicolumn{8}{l}{\textbf{Benchmark}: AgentBench-Operating System} \\
\midrule
GPT-4.1-Nano & 0.307 & 0.725 & 0.756 & 0.768 & 0.770 & 0.757 & \underline{0.779} & 0.775 & \textbf{0.781} \\
GPT-3.5-Turbo & 0.275 & 0.747 & 0.750 & \underline{0.782} & 0.765 & 0.765 & 0.749 & 0.777 & \textbf{0.791} \\
Gemma-2-27b-it & 0.289 & 0.747 & 0.636 & \underline{0.760} & 0.755 & 0.652 & 0.766 & 0.755 & \textbf{0.814} \\
DeepSeek-V3 & 0.310 & \underline{0.729} & 0.636 & 0.724 & 0.716 & 0.655 & 0.717 & 0.722 & \textbf{0.767} \\
Qwen2.5-72B-Instruct & 0.508 & 0.625 & 0.620 & \textbf{0.707} & 0.687 & 0.631 & 0.678 & 0.678 & \underline{0.704} \\
\midrule
\multicolumn{1}{c}{\textbf{Average}} & 0.338 & 0.715 & 0.679 & \underline{0.748} & 0.738 & 0.692 & 0.738 & 0.741 & \textbf{0.771} \\

\midrule
\midrule
& & \multicolumn{8}{l}{\textbf{Benchmark}: StrategyQA} \\
\midrule

GPT-4.1-Nano & 0.691 & 0.512 & 0.492 & \underline{0.542} & 0.503 & 0.502 & 0.499 & 0.527 & \textbf{0.544} \\
GPT-3.5-Turbo & 0.611 & 0.593 & 0.438 & 0.623 & \textbf{0.611} & 0.440 & 0.600 & 0.607 & \underline{0.604} \\
Gemma-2-27b-it & 0.777 & \underline{0.698} & 0.615 & 0.669 & 0.624 & 0.622 & 0.640 & 0.667 & \textbf{0.766} \\
DeepSeek-V3 & 0.790 & 0.573 & 0.548 & 0.559 & 0.558 & \underline{0.575} & 0.574 & 0.563 & \textbf{0.607} \\
Qwen2.5-72B-Instruct & 0.796 & 0.500 & 0.495 & \underline{0.573} & \underline{0.573} & 0.493 & 0.567 & 0.563 & \textbf{0.617} \\
\midrule
\multicolumn{1}{c}{\textbf{Average}} & 0.733 & 0.575 & 0.518 & \underline{0.593} & 0.574 & 0.526 & 0.576 & 0.585 & \textbf{0.628} \\

\bottomrule
\end{tabular}
    }
    \vspace{-2mm}
\end{table*}

\begin{figure*}
    \centering
    \includegraphics[width=1\linewidth]{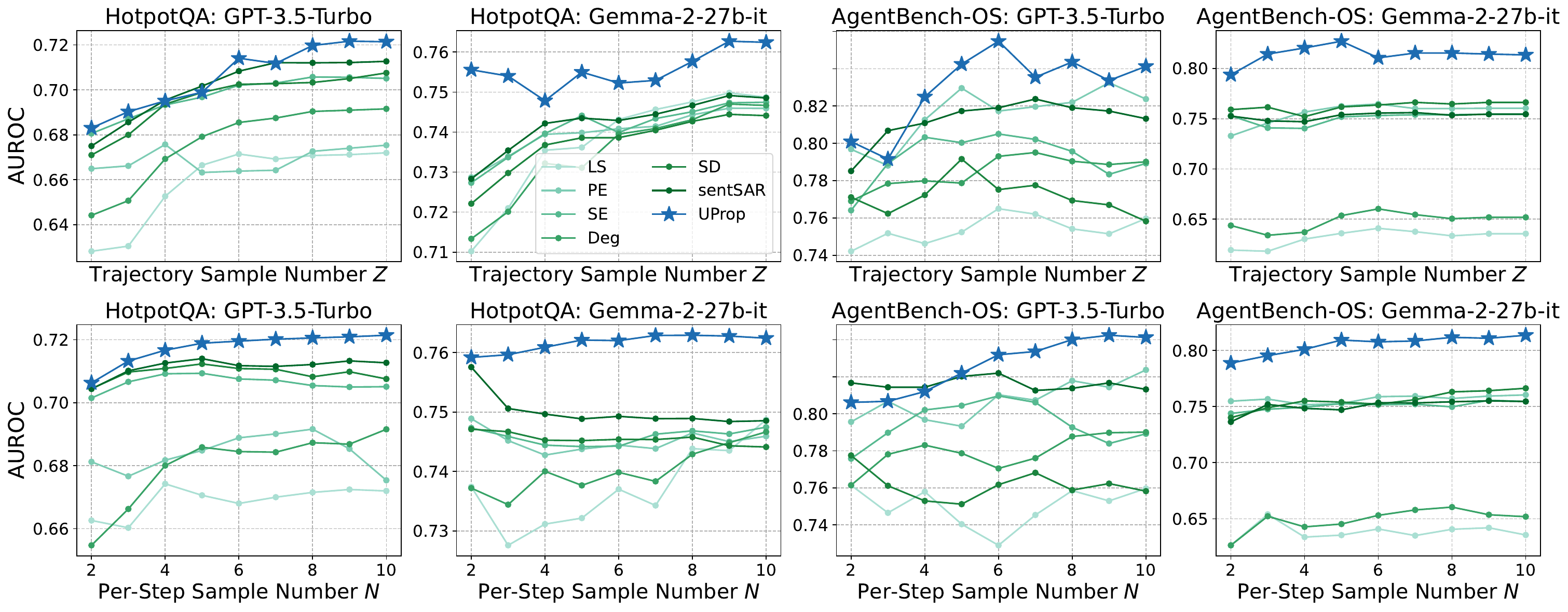}
    \caption{Comparing the sampling efficiency of \methodname with baselines.}
    \label{fig:ablation_sample_efficiency}
    \vspace{-5mm}
\end{figure*}

\textbf{LLMs and Sampling} \,\, We consider state-of-the-art commercial LLMs (GPT-4.1-Nano-2025-04-14~\cite{achiam2023gpt}, GPT-3.5-Turbo-0125~\cite{brown2020language}) and open-source LLMs (QWen2.5-72b-Instruct~\cite{yang2024qwen2}, DeepSeek-V3~\cite{DeepSeekAI2024DeepSeekV3TR}, and Gemma-2-27b-it~\cite{Riviere2024Gemma2I}) as backbones. For generative hyper-parameters, we use greedy search to generate responses for correctness evaluation and multinomial search with temperature set to 0.8 for MC sampling. For all the generations, we set the maximum number of new tokens to be 512. By default, the trajectory sample number, i.e., $Z$, and the per-step sample number, i.e., $N$, are set to 10.

\textbf{Baselines} \,\, We consider 7 popular single-step LLM UQ methods: Perplexity (PPL), Lexical Similarity (LS)~\cite{Fomicheva2020UnsupervisedQE}, PE~\cite{malinin2020uncertainty}, SE~\cite{kuhnsemantic}, Deg~\cite{lingenerating}, SD~\cite{qiu2024semantic}, sentSAR~\cite{duan2024shifting}. 
For a fair comparison and also a straightforward adaptation, baselines are calculated over the same TDP samples. Specifically, 
for each TDP sample, baseline methods first \ding{202} calculate per-step uncertainty by their design. Then, \ding{203} the TDP total uncertainty is aggregating all these per-step uncertainties by either \textit{average} or \textit{Root Mean Square (RMS)}. Eventually, \ding{204} the final uncertainty is the averaging of TDPs' total uncertainties.

\textbf{Evaluation Metric} \,\, Following existing work~\cite{kuhnsemantic} in this domain, we evaluate UQ by assessing how well it predicts the correctness of the model’s generated answers for a given question, with the metric Area Under the Receiver Operating Characteristic Curve (AUROC).

\subsection{Performance on Multi-Step Decision-Making Benchmarks}
We report the general performance (Success Rate) of LLMs and the AUROC of baselines and \methodname, over AgentBench-Operating System (OS) and StrategyQA datasets. The performance of baselines aggregated by \textit{averaging} is reported in~\cref{tab:average_baseline} (please refer to~\cref{appendix:rms_agg} for RMS aggregation comparison). It is shown that \methodname achieves the best UQ performance in most settings, compared to both average and RMS aggregation. It significantly outperforms existing methods in general, e.g., \methodname outperforms baselines by 2.3\% $\sim$ 9.2\%  AUROC in AgentBench-OS and 3.5\% $\sim$ 11\% AUROC in StrategyQA.

\subsection{Sampling Efficiency}

To quantify the sampling efficiency, we compare the AUROC of sampling-based baselines and \methodname over various TDP sampling numbers, i.e., $Z\in[2, 10]$, and per-step sampling numbers within each TDP, i.e., $N \in [2, 10]$ (we only vary one of the sampling numbers at each time and fix the other sampling numbers to be 10). Results are summarized in~\cref{fig:ablation_sample_efficiency}. It is shown that \methodname outperforms baselines in most sampling configurations, including when very few trajectory samplings or per-step samplings are available. It implies that \methodname is effective and efficient in LLM multi-step UQ.

\begin{wraptable}{r}{7cm}
\vspace{-12mm}
    \centering
    \caption{Ablation study of IU and EU in \methodname.}
    \label{tab:uprop_ablation}
    \adjustbox{width=0.5\textwidth}{
    \begin{tabular}{lcccc}
    \toprule
      \textbf{Model} & \methodname & w/o \texttt{EU} & w/o \texttt{IU} \\
      \toprule
      GPT-4.1-Nano & \textbf{0.781} & 0.726 (\textbf{-5.5\%}) & 0.770 (-1.1\%) \\
      GPT-3.5-Turbo & \textbf{0.791} & 0.747 (-4.4\%) & 0.717 (\textbf{-7.4\%}) \\
      Gemma-2-27b-it & \textbf{0.813} & 0.765 (\textbf{-4.8\%}) & 0.794 (-1.9\%) \\
      DeepSeek-V3 & \textbf{0.767} & 0.700 (\textbf{-6.7\%}) & 0.733 (-3.4\%) \\
      Qwen2.5-72B-Instruct & \textbf{0.704} & 0.652 (\textbf{-5.2\%}) & 0.684 (-2.0\%) \\
      \bottomrule
    \end{tabular}
    }
    \vspace{-3mm}
\end{wraptable}

\subsection{Ablation Study}
\textbf{IU vs. EU} \,\, We investigate the effectiveness of \texttt{IU} and \texttt{EU} individually.
In~\cref{tab:uprop_ablation}, we provide the AUROC when removing each of these components from \methodname, over the AgentBench-OS benchmark.
In general, both \texttt{IU} and \texttt{EU} contribute to the performance improvement. However,  w/o \texttt{EU} brings larger performance drops than IU, indicating that EU is an essential component in uncertainty quantification.

\begin{wraptable}{r}{9cm}
    \centering
    \vspace{-4mm}
    \caption{The evaluation of selective prediction with AUARC.}\label{tab:ablation_auarc}
    \adjustbox{width=0.65\textwidth}{
    \begin{tabular}{lccccccc|c}
\toprule
\textbf{Models} & PPL & LS & PE & SE & Deg & sentSAR & SD & \methodname \\
\midrule
GPT-4.1-Nano & \underline{67.2} & 62.9 & 68.2 & 66.6 & 63.8 & 67.2 & 66.7 & \textbf{68.5} \\
GPT-3.5-Turbo & 64.1 & 54.0 & \textbf{67.2} & 65.1 & 54.1 & 64.6 & 64.6 & \underline{66.8} \\
Gemma-2-27b-it & \underline{84.5} & 79.7 & 83.2 & 81.8 & 80.2 & 83.1 & 82.2 & \textbf{86.0} \\
DeepSeek-V3 & 77.5 & 76.1 & \underline{78.6} & 78.4 & 77.9 & \underline{78.6} & 79.2 & \textbf{79.7} \\
Qwen2.5-72B-Instruct & 74.1 & 75.7 & 78.4 & \underline{78.5} & 76.2 & 78.2 & 77.5 & \textbf{81.1} \\
\midrule
\textbf{Average} & 73.5 & 69.7 & \underline{75.1} & 74.1 & 70.4 & 74.3 & 74.0 & \textbf{76.4} \\
\bottomrule
\end{tabular}
    }
    \vspace{-4mm}
\end{wraptable}

\textbf{Selective Prediction} \,\, Rejecting response by uncertainty is an important UQ application, e.g., hallucination detection in LLMs. In~\cref{tab:ablation_auarc}, we report the selective prediction performance comparison, evaluated by metric Area Under Accuracy-Rejection Curve (AUARC)~\cite{Nadeem2009AccuracyRejectionC} over the StrategyQA benchmark. We show that \methodname substantially outperforms baselines in most cases, e.g., \methodname outperforms baselines by up to 2.6\% AUARC. This indicates that \methodname retains better performance in rejecting incorrect answers. 

\begin{wrapfigure}{r}{7cm}
\centering
    \includegraphics[width=\linewidth]{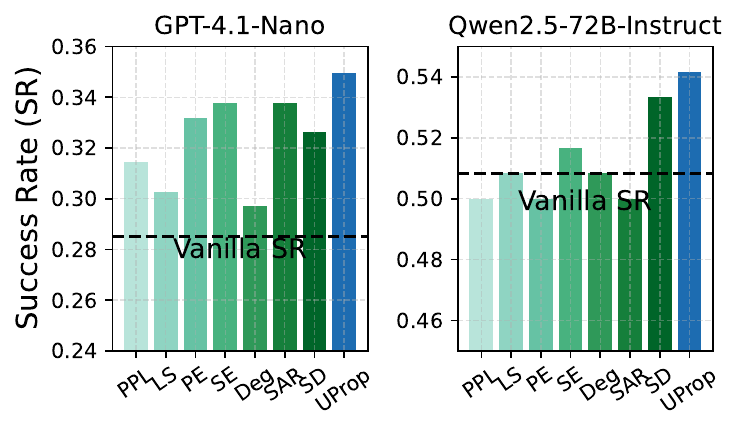}
    \caption{Uncertainty as the correctness indicator for improved LLM performance.}
    \label{fig:ablation_tts}
    \vspace{-5mm}
\end{wrapfigure}

\textbf{Uncertainty as Correctness Indicator} \,\, Uncertainty could serve as the indicator of correctness, which is one of the potential applications of UQ. We study the effectiveness of baselines and \methodname in identifying correct answers from multiple generations. Specifically, for each question, we first sample 10 generations and then select the one with the lowest uncertainty (estimated by various UQ methods) as the final answer. We calculate the general performance, i.e., success rate (SR), of these final answers. We conduct this experiment over AgentBench-OS and the results are reported in~\cref{fig:ablation_tts}. We show that UQ effectively improves SR and \methodname achieves the best performance among all the baselines.

\subsection{Intermediate Uncertainty Propagation Analysis}
To understand how uncertainty is propagated along decision trajectories and identify the contributions of IU and EU individually, we provide the uncertainty percentage of IU and EU, i.e., $\frac{IU}{IU+EU}$ and $\frac{EU}{IU+EU}$, at each decision step. Results are summarized in~\cref{fig:uncertainty_percentage} (the detailed per-model results are provided in~\cref{appendix:uncertainty_percentage}). We observe that \ding{202} IU usually contributes significantly to the first few decision steps while EU dominates the rest. As the decision step grows, EU heavily affects the total uncertainty of the decision step, highlighting the significance of EU in uncertainty propagation; \ding{203} GPT-4.1-Nano embraces a relatively smaller EU percentage compared to other LLMs at the later decision steps, e.g., EU and IU share closer percentages at the end step in StrategyQA. This indicates that GPT-4.1-Nano has more stable and less uncertain decisions.

\begin{figure*}[t]
    \centering
    \includegraphics[width=\linewidth]{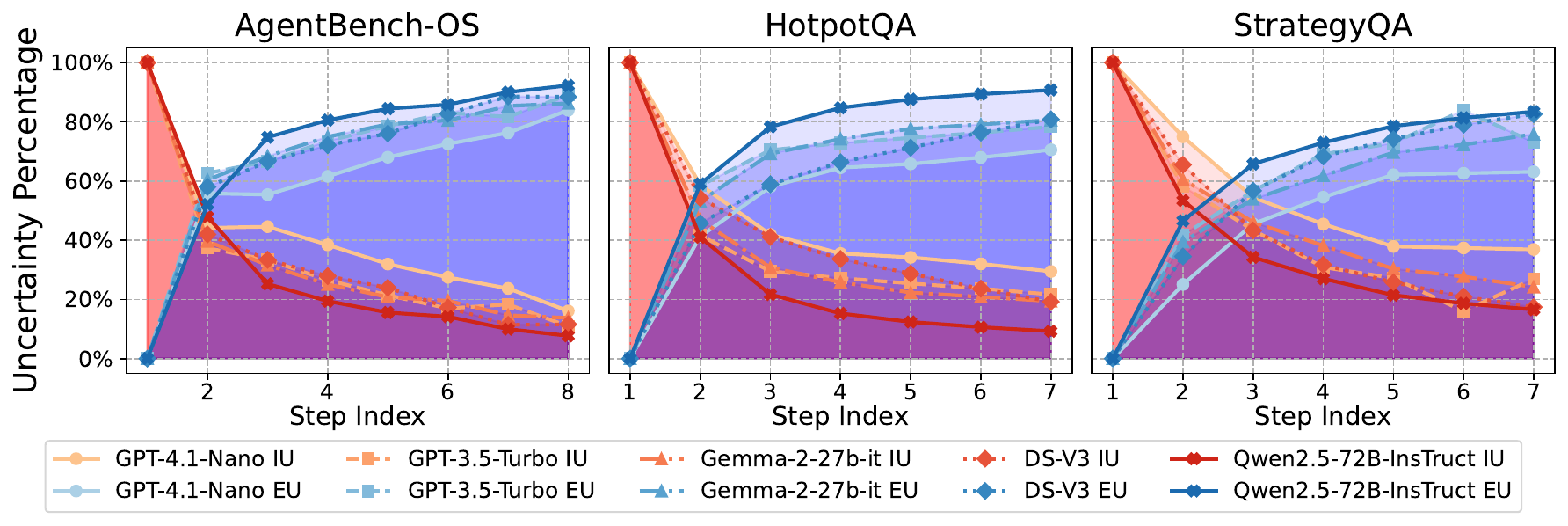}
    \caption{The percentage of intrinsic and extrinsic uncertainty at each step. The Red shadow area is the percentage of IU and the Blue shadow area is the percentage of EU.}
    \label{fig:uncertainty_percentage}
    \vspace{-5mm}
\end{figure*}

\section{Related Work}

\textbf{Uncertainty Quantification (UQ) of LLMs} \,\, In auto-regressive LLMs, UQ quantifies the uncertainty within its prediction distribution~\cite{malinin2020uncertainty}. From the perspective of entropy, uncertainty could be measured by the log probability of generations sampled from the output space~\cite{gawlikowski2023survey}. However, entropy may overestimate uncertainty due to the semantic clusters. To address this issue, Semantic Entropy (SE)~\cite{kuhnsemantic} clusters LLM outputs by semantics and then calculates cluster-wise entropy as the uncertainty. Deg~\cite{lingenerating} is specifically designed for black-box UQ and it models output consistency by either node connectivity or eigenvalues of a semantic graph (which is further extended by INSIDE~\cite{chen2024inside} in LLM hidden space). SAR~\cite{duan2024shifting} reveals token-level and sentence-level semantic imbalance in LLM UQ. The token-level semantic importance is further extended by CSL~\cite{lin2024contextualized}. Semantic Density (SD)~\cite{qiu2024semantic} draws the idea from maximum probability and calculates the density of a target generation within a semantic space as the uncertainty. To mitigate the ``semantic overlapping'' among semantic clusters, KLE~\cite{nikitin2024kernel} encodes semantic similarities of LLM outputs via positive semidefinite unit trace kernels and quantifies uncertainty using the von Neumann entropy.

\textbf{LLM Multi-Step Decision-making} \,\, LLM Multi-step decision-making refers to a sequential decision-making process~\cite{wang2024survey}, where a LLM agent continuously interacts with an environment. Out of various purposes, the realizations of the environment can be diverse such as Operating System (OS)~\cite{liu2023agentbench}, Wikipedia, Games~\cite{duan2024gtbench}, and Robotics~\cite{liu2024aligning}. Moreover, multi-step decision-making also involves improved LLM reasoning for better query performance, e.g., question answering. ReAct~\cite{yao2023react} proposes the ``thinking-action-observation'' paradigm where LLM submits sequential actions for answering questions. Reflection~\cite{shinn2023reflexion} and Rest~\cite{aksitov2023rest} incorporate self-reflection~\cite{ji2023towards} in the ReAct framework for improved multi-step reasoning. Q$^*$~\cite{wang2024q} includes deliberating planning for multi-step LLM reasoning. Many work also involves auxiliary modules, such as Graph~\cite{wu2025graph} and Tool~\cite{paranjape2023art}, to enhance multi-step logical reasoning. \cite{wang2025stepwise} improves LLM reasoning by proactively referencing underutilized information and minimizing redundant information.

\section{Conclusion}
In this paper, we investigate the uncertainty propagation of LLMs in multi-step decision-making. Specifically, we first provide a principled, information-theoretical framework that decomposes the uncertainty into intrinsic uncertainty and extrinsic uncertainty. We then propose \methodname, as an efficient and effective estimator of extrinsic uncertainty. We conduct experiments over popular sequential decision-making scenarios and experimental results demonstrate the superior performance of \methodname compared to best-performing baselines.

\textbf{Limitations \& Social Impacts} \,\, 
The proposed \methodname relies on MC sampling for MI estimation. On the one hand, the estimation might be deviated due to insufficient sampling and unknown distribution from the LLM decision space. Moreover, sampling may result in latency in real-world deployment. Also, our study involves closed-source commercial LLMs such as GPT-4.1 and GPT-3.5-Turbo, which may suffer from reproducibility issues due to the continuous updating of these models.

We investigate the uncertainty quantification in LLMs, which is one of the most important topics in trustworthy LLMs and responsible LLMs. We expect our method will improve hallucination detection in LLM sequential decision-making and could be used to correct LLM behaviors in uncertain decision scenarios.

\section*{Acknowledgement}
This work was performed under the auspices of the U.S. Department of Energy by Lawrence Livermore National Laboratory under Contract DE-AC52-07NA27344 and LDRD
Program Project No.~23-ERD-030 (LLNL-JRNL-2006778). SM was supported by the U.S. Department of Energy, Office of Science, Advanced Scientific Computing Research, through the SciDAC-RAPIDS2 institute under Contract DE-AC02-06CH11357. 

{
    \bibliographystyle{ieeenat_fullname}
    \bibliography{main}
}

\clearpage

\appendix

\section{Implementation Details and Related Proofs}

In this section, we provide detailed proofs and procedures used in this paper. We will reuse the notations we defined before.

\subsection{Uncertainty Decomposition}\label{appendix:uncertainty_decompose}
The joint distribution for a sequence of events $\vy_{1:t}$, conditioned on $x$, follows the chain rule for conditional probability:

\begin{equation}
    \begin{aligned}
        p_{\vtheta}(\vy_{1:t}|\vx) = p_{\vtheta}(\vy_1|\vx)p_{\vtheta}(\vy_2|\vy, 1\vx)p_{\vtheta}(\vy_3|\vy_{1:2}\vx) \cdots p_{\vtheta}(\vy_t|\vy_{1:t-1} \vx).
    \end{aligned}
\end{equation}
By marginalizing out previous decisions $\vy_{1:t-1}$, the distribution $p_{\vtheta}(\vy_{t}|\vx)$ becomes:
\begin{equation}
\begin{aligned}
    p_{\vtheta}(\vy_{t}|\vx) & = \int{ p_{\vtheta}(\vy_t|\vy_{1:t-1}, \vx)  p_{\vtheta}(\vy_{1:t-1}|\vx) d\vy_{1:t-1} } \\
    & = \int{p_{\vtheta}{(\vy_t|\vy_{1:t-1}, \vx})}\prod\nolimits_{i}^{t-1}{p_{\vtheta}(\vy_i|\vy_{1:i-1}, \vx)}d\vy_1 d\vy_2 \cdots d\vy_{i-1}
\end{aligned}
\end{equation}

\subsection{Entropy Decomposition}\label{appendix:entropy_decom}
Based on conditional mutual information and iteratively apply it to each preceding decision $\vy_i$, we have the following decomposition:

\begin{equation}
    \begin{aligned}
        H(\vy_t|\vx) & = H(\vy_t|\vy_{t-1}, \vx) + I(\vy_t;\vy_{t-1}|\vx) \\
        & = H(\vy_t|\vy_{t-2:t-1}, \vx) + I(\vy_t;\vy_{t-2}|\vy_{t-1}, \vx) + I(\vy_t;\vy_{t-1}|\vx) \\
        & \cdots \\
        & = H(\vy_t|\vy_{1:t-1}, \vx) + \sum\nolimits_{i}^{t-1} {I(\vy_t;\vy_i|\vy_{i+1:t-1}, \vx)}.
    \end{aligned}
\end{equation}

\subsection{MI Calculation Necessities Exponential Exploration}\label{appendix:exponential_explore}
In a multi-step decision-making process, we denote by $\mathcal{A}$ the decision space at each step, the MI between the $n$-th step distribution $\vy_{n}$ and the $m$-th step distribution $\vy_m$ $(m>n)$, i.e., $I(\vy_m;\vy_n)$ requires the joint distribution $p_{\vtheta}(\vy_m, \vy_n|\vx)$:
\begin{equation}
    p_{\vtheta}(\vy_m, \vy_n|\vx) = \idotsint_{\mathcal{A}^{m-n-1}} {p_{\vtheta} (\vy_1, \vy_2, \cdots, \vy_t, \cdots, \vy_m | \vx)} \prod_{k \in \{1,\cdots,m\} \setminus \{n.m\}} {d\vy_k}.
\end{equation}
Each of the ($\Delta=m-n-1$) intermediate steps introduces an independent integral over the entire action domain $\mathcal{A}$ turning the calculation into an $\Delta$-fold (hyper-)integral whose effective cost grows as $\mathcal{O}(|\mathcal{A}|^\Delta)$. Thus, the volume of the decision sub-space expands exponentially with the gap $\Delta$.

\subsection{Proof of Theorem 1: Convergence of the TDP sampling}\label{appendix:proof_MI_converge}

Given a TDP $z$, based on~\cref{eq:uprop_approx}, the total uncertainty at step $t$ is:
\begin{equation}\label{eq:g_t_z}
    \hat{g}_t(z) = H(\vy_t|\vy_{1:t-1}, \vx) + \sum_{i}^{t-1} \widehat{\textit{PMI}}(\vy_t;\vy_i^{(k)}|\vy_{i+1:t-1}^{(k)}, \vx)\Big),
\end{equation}
Taking the expectation of~\cref{eq:g_t_z}, we obtain $\mathbb{E}_{z\sim\mathcal{Z}}[\hat{g}_t(z)] = H(\vy_t|\vx)$.
With independent TDPs sampled $\mathcal{Z} = \{z_1, z_2, \cdots\}$, then
$$
\hat{H}(\mathcal{P}_\textit{TDP}) = \frac{1}{|\mathcal{Z}|}\sum_z^Z \sum_t^{T_z} \hat{g}_t(z),
$$
where $T_z$ is the length of the trajectory $z$. Similarly, taking the expectation over $\mathcal{Z}$, we obtain
\begin{equation}
    \mathbb{E}_{z\sim\mathcal{Z}} [\hat{H}(\mathcal{P}_\textit{TDP})] = \sum_t^T H(\vy_t|\vx) = H(\mathcal{P}).
\end{equation}
Thus, we show the estimator is unbiased. With the law of large numbers, we have
$$
\hat{H}(\mathcal{P}_\textit{TDP}) = \frac{1}{|\mathcal{Z}|}\sum_z^Z G(z) \rightarrow \mathbb{E}_{z\sim\mathcal{Z}}[G(z)] = H(\mathcal{P}),
$$
with $N \rightarrow \infty$, where $G(z) = \sum_t^{T_z} \hat{g}_t(z)$.

\subsection{Proof of Theorem 2: Convergence of the PMI Approximation}\label{appendix:proof_PMI_estimation}
We start from the definition of the PMI conditioned on $\vx$: $\textit{PMI}(\vy_t;\vy_{t-1}^{(k)} | \vx) = \log \frac{p(\vy_t|\vy_{t-1}^{(k)}, \vx)}{p(\vy_t|\vx)}$, where $\vy_{t-1}^{(k)}$ is a realization. According to the local smoothness assumption, for any given $\vy_{t-1}^{\prime}$ sufficiently close to the given $\vy_{t-1}^{(k)}$, it must hold that $p(\vy_t|\vy_{t-1}^{\prime}, \vx) \approx  p(\vy_t|\vy_{t-1}^{(k)}, \vx)$. Consider the marginalization over $\vy_{t-1}$:
\begin{equation}
\begin{aligned}
    p(\vy_t|\vx) & = \int p(\vy_t|\vy_{t-1}^{\prime}, \vx)p(\vy_{t-1}^\prime|\vx) d\vy_{t-1}^\prime. \\
\end{aligned}
\end{equation}
Under the smoothness assumption, within the kernel radius around $\vy_{t-1}^{(k)}$, we can write:
\begin{equation}
\begin{aligned}
    p(\vy_t|\vx) & \approx \int p(\vy_t|\vy_{t-1}^{(k)}, \vx)K_\tau(dist(\vy_{y_{t-1}}, \vy_{t-1}^{(k)}))p(\vy_{t-1}|\vx)d\vy_{t-1} \,\, \text{\ding{226} \textit{Local Smoothness Assumption}} \\
    & = p(\vy_t|\vy_{t-1}^{(k)}, \vx) \int K_\tau(dist(\vy_{y_{t-1}}, \vy_{t-1}^{(k)}))p(\vy_{t-1}|\vx)d\vy_{t-1} \\
    & \approx p(\vy_t|\vy_{t-1}^{(k)}, \vx) \sum_i^N K_\tau(dist(\vy_{t-1}^{(i)}, \vy_{t-1}^{(k)})) \,\, \text{\ding{226} \textit{MC Approximation}} \\
    & = \hat{p}(\vy_t|\vx)
\end{aligned}
\end{equation}
It is shown that as the sampling number $N \rightarrow \infty$, $\hat{p}(\vy_t|\vx) \rightarrow p(\vy_t|\vx)$, thus $\widehat{\textit{PMI}}(\vy_t|\vy_{t-1}, \vx) \rightarrow \textit{PMI}(\vy_t|\vy_{t-1}, \vx)$.

\subsection{Measuring Distance between LLM Agentic Decisions}\label{apendix:LLM_decision_dist}

In our decision-making environments, at each decision step, LLMs are prompted to provide a \textbf{\textit{Reasoning}} output, then followed by an \textbf{\textit{Action}}. Though the \textit{Reasoning} output is long and versatile, the generated \textit{Action} is usually pre-defined to be short and concise, such as \texttt{SEARCH(<keyword>)} and \texttt{LOOKUP(<keyword>)} in the ReAct agent. Moreover, considering the decision is largely represented by \textit{Action}, the distance between \textit{Actions} becomes an effective measurement of the decision distance. In this way, string fuzzy matching is an efficient method to measure the distance between short actions. Existing work usually applies auxiliary models such as Natural Language Inference (NLI)~\cite{kuhnsemantic} model and embedding models~\cite{duan2024shifting}. In multi-step decision-making, due to the significantly larger query numbers than single-turn QA, these models will be costly and infeasible in the real world.

\section{Experimental Setup}

\subsection{AgentBench-Operating System Benchmark}\label{appendix:agentbench_os}

The Operating System (OS) split in AgentBench~\cite{liu2023agentbench} enables LLMs to interact with and operate within real operating system (OS) environments through terminal commands presents both an exciting opportunity and a significant challenge. It evaluates LLMs in genuine bash environments (specifically, Ubuntu Docker containers using a suite of human-written tasks. These tasks require either deterministic responses—such as querying the number of users without home directories—or the execution of command sequences to achieve practical objectives—such as recursively setting all directory files to read-only, excluding specific users. In this split, there are 144 questions in the testing set, with the expected interaction turns to be 1200.

\textbf{Evaluation Metric}: The performance is evaluated by \textbf{\textit{Success Rate}} (SR): a task is deemed as a success if the LLM returned answer matches the ground-truth label.

We follow the AgentBench codebase\footnote{https://github.com/THUDM/AgentBench} and use the official one-shot demonstration for each OS task. The one-shot demonstration and one test task demonstration is provided as the following:
\begin{tcolorbox}[fontupper=\small, title={One-Shot Demonstration}]
\textbf{<USER>}

You are an assistant that will act like a person, I'will play the role of linux(ubuntu) operating system. Your goal is to implement the operations required by me or answer to the question proposed by me. For each of your turn, you should first think what you should do, and then take exact one of the three actions: "bash", "finish" or "answer". 

1. If you think you should execute some bash code, take bash action, and you should print like this:

Think: put your thought here.

Act: bash

```bash
\# put your bash code here
```

2. If you think you have finished the task, take finish action, and you should print like this:

Think: put your thought here.

Act: finish

3. If you think you have got the answer to the question, take answer action, and you should print like this:

Think: put your thought here.

Act: answer(Your answer to the question should be put in this pair of parentheses)

If the output is too long, I will truncate it. The truncated output is not complete. You have to deal with the truncating problem by yourself. Attention, your bash code should not contain any input operation. Once again, you should take only exact one of the three actions in each turn.

Now, my problem is:

tell me how many files are in the directory "/etc"?

\textbf{<LLM Agent>}

Think: To count the files in /etc, I need to print all the files in it.

Act: bash

```bash
ls /etc
```

\textbf{<USER>}

The output of the OS:
cpi cron.hourly fuse.conf iproute2 lvm networkd-dispatcher protocols selinux tmpfiles.d [truncated because the output is too long]

\textbf{<LLM Agent>}

Think: The output has been truncated because it is too long, so I need to count files by script directly.

Act: bash

```bash
ls -1 /etc | wc -l
```

\textbf{<USER>}

The output of the OS:
220

\textbf{<LLM Agent>}

agent Think: Now I get the answer, it is 220.

Act: answer(220)

\textbf{<USER>}

Now, I will start a new problem in a new OS. My problem is: \textbf{\textit{<Test Question>}}

\end{tcolorbox}

\begin{tcolorbox}[fontupper=\small, title={Test Task Demonstration}]

\textbf{<USER>}

{Tell me how many files can be executed in PATH.
}

\textbf{<LLM Agent>}

Think: To count the executable files in PATH, I need to list all directories in PATH and count the executable files in each directory.

Act: bash

```bash
for dir in \$(echo \$PATH | tr ':' ' '); do
    echo -n "\$dir: "
    find \$dir -maxdepth 1 -type f -executable | wc -l
done
```

\textbf{<USER>}

The output of the OS:

    echo -n "\$dir: "
>     find \$dir -maxdepth 1 -type f -executable | wc -l
> done
/usr/local/sbin: 0
/usr/local/bin: 0
/usr/sbin: 109
/usr/bin: 446
/sbin: 0
/bin: 0

\textbf{<LLM Agent>}

Think: The total number of executable files in PATH is the sum of the counts in each directory.

Act: bash

```bash
echo \$((0 + 0 + 109 + 446 + 0 + 0))
```
user The output of the OS is empty.
agent Think: The total number of executable files in PATH is 555.

Act: answer(555)
\end{tcolorbox}

\subsection{Multi-Step Reasoning: HotpotQA and StrategyQA}\label{appendix:multihop_react}

\textbf{HotpotQA} is a question-answering dataset featuring multi-hop questions: each question needs support from multiple facts. For instance, the question ``\textit{What country of origin does House of Cosbys and Bill Cosby have in common?}'' requires facts regarding ``House of Cosbys'' and ``Bill Cosby''. We follow the ReAct~\cite{yao2023react} experimental setting and randomly select 500 questions from the test split and prompt LLMs in a ``Reasoning-Action-Observation'' manner:

\begin{tcolorbox}[fontupper=\small, title=\textbf{Prompt Template}]
Solve a question answering task with interleaving Thought, Action, Observation steps. Thought can reason about the current situation, and Action can be three types: 
    (1) Search[entity], which searches the exact entity on Wikipedia and returns the first paragraph if it exists. If not, it will return some similar entities to search.
    (2) Lookup[keyword], which returns the next sentence containing keyword in the current passage.
    (3) Finish[answer], which returns the answer and finishes the task.
    Here are some examples.
    \textbf{\textit{<Few-Shot Demonstration>}}
\end{tcolorbox}

\begin{tcolorbox}[fontupper=\small, title=\textbf{Few-Shot Demonstrations (HotpotQA)}]
\textbf{Question}: What is the elevation range for the area that the eastern sector of the Colorado orogeny extends into?

\textbf{Thought 1}: I need to search Colorado orogeny, find the area that the eastern sector of the Colorado orogeny extends into, then find the elevation range of the area.

\textbf{Action 1}: Search[Colorado progeny]

\textbf{Observation 1}: The Colorado orogeny was an episode of mountain building (an orogeny) in Colorado and surrounding areas.

\textbf{Thought 2}: It does not mention the eastern sector. So I need to look up eastern sector.

\textbf{Action 2}: Lookup[eastern sector]

\textbf{Observation 2}: (Result 1 / 1) The eastern sector extends into the High Plains and is called the Central Plains progeny.

\textbf{Thought 3}: The eastern sector of Colorado orogeny extends into the High Plains. So I need to search High Plains and find its elevation range.

\textbf{Action 3}: Search[High Plains]

\textbf{Observation 3}: High Plains refers to one of two distinct land regions:

\textbf{Thought 4}: I need to instead search High Plains (United States).

\textbf{Action 4}: Search[High Plains (United States)]

\textbf{Observation 4}: The High Plains are a subregion of the Great Plains. From east to west, the High Plains rise in elevation from around 1,800 to 7,000 ft (550 to 2,130 m).[3]

\textbf{Thought 5}: High Plains rise in elevation from around 1,800 to 7,000 ft, the answer is 1,800 to 7,000 ft.

\textbf{Action 5}: Finish[1,800 to 7,000 ft]
\end{tcolorbox}

\textbf{StrategyQA} is also a popular multi-hop reasoning benchmark, which requires steps for answering the question. 
It includes 2,780 examples, with a strategy question, decompositions, and evidence paragraphs within each example. Similar to HotpotQA, we also randomly select 200 questions and prompts LLMs in the ReAct manner with the following few-shot demonstrations.

\begin{tcolorbox}[fontupper=\small, title=\textbf{Few-Shot Demonstrations (StrategyQA)}]

\textbf{Question}: Is Mixed martial arts totally original from Roman Colosseum games?

\textbf{Thought 1}: Mixed martial arts (MMA) does have some similarities to the ancient Roman games held in the Colosseum, where gladiators would fight to the death as a form of entertainment. However, there are also distinct differences between the two, such as rules, regulations, and cultural contexts.

\textbf{Action 1}: Search[Roman Colosseum games]

\textbf{Observation 1}: The Roman Colosseum, also known as the Flavian Amphitheater, was a huge arena used for gladiatorial contests, animal hunts, executions, re-enactments of famous battles, and dramas. The events held at the Colosseum were brutal displays of power and entertainment for the ancient Romans.

\textbf{Thought 2}: I need to further check mixed martial arts.

\textbf{Action 2}: Lookup[Mixed martial arts]

\textbf{Observation 2}: Mixed martial arts is a full-contact combat sport that allows a wide variety of fighting techniques and skills from a mixture of other combat sports to be used in competition. It involves striking and grappling techniques, both standing and on the ground.

\textbf{Thought 3}: While both the Roman Colosseum games and modern mixed martial arts involve combat sports, MMA is a regulated sport with rules and safety measures that were not present in the ancient gladiatorial contests. The styles of fighting and the purposes of the two are different, with MMA focusing more on competitive sportsmanship rather than the spectacle of bloodshed and death seen in the Roman games.

\textbf{Action 3}: Finish[No]

\end{tcolorbox}

For both HotpotQA and StrategyQA, we use the Wikipedia engine and the environment that takes the retrieval command from LLM Agents and returns the required information.

\section{Extended Experimental Results}

\subsection{Aggregating Baselines with RMS}\label{appendix:rms_agg}

In~\cref{tab:rms_baseline}, we compare \methodname with baselines aggregated by Rooted Mean Square (RMS). RMS aggregation is mainly used to address ``outlier'' trajectories such as exceptionally large steps and/or large uncertainties. It is shown that RMS aggregation is worse and simple averaging (\cref{tab:average_baseline}) and \methodname is significantly better than it.

\begin{table*}[h]
    \centering
    \caption{AUROC results over AgentBench-Operating System and StrategyQA benchmarks. For single-turn baseline UQ methods, uncertainties are aggregated by \textbf{\textit{RMS}} over all steps.}
    \label{tab:rms_baseline}
    \adjustbox{width=\textwidth}{
    \begin{tabular}{lcccccccc|c}
\toprule
\multicolumn{1}{c}{\textbf{Models}} & \textbf{Success Rate} & PPL & LS & PE & SE & Deg & SD & sentSAR & \methodname (ours) \\
\midrule
& & \multicolumn{8}{l}{\textbf{Benchmark}: AgentBench-Operating System} \\
\midrule
GPT-4.1-Nano & 0.307 & 0.710 & 0.761 & 0.768 & 0.754 & 0.762 & 0.765 & \underline{0.769} & \textbf{0.781} \\
GPT-3.5-Turbo & 0.275 & 0.722 & 0.739 & 0.772 & 0.756 & 0.752 & 0.739 & \underline{0.774} & \textbf{0.791} \\
Gemma-2-27b-it & 0.289 & 0.731 & 0.639 & 0.750 & 0.739 & 0.653 & \underline{0.755} & 0.754 & \textbf{0.814} \\
DeepSeek-V3 & 0.310 & 0.704 & 0.621 & \underline{0.711} & 0.693 & 0.631 & 0.691 & 0.705 & \textbf{0.767} \\
Qwen2.5-72B-Instruct & 0.508 & 0.604 & 0.614 & \underline{0.695} & 0.668 & 0.627 & 0.644 & 0.641 & \textbf{0.704} \\
\midrule
\multicolumn{1}{c}{\textbf{Average}} & 0.338 & 0.694 & 0.675 & 0.739 & 0.722 & 0.685 & 0.719 & \underline{0.729} & \textbf{0.771} \\

\midrule
\midrule
& & \multicolumn{8}{l}{\textbf{Benchmark}: StrategyQA} \\
\midrule

GPT-4.1-Nano & 0.691 & 0.516 & 0.505 & \textbf{0.551} & 0.506 & 0.520 & 0.502 & 0.539 & \underline{0.544} \\
GPT-3.5-Turbo & 0.611 & 0.607 & 0.435 & \textbf{0.620} & 0.608 & 0.438 & 0.601 & 0.530 & \underline{0.604} \\
Gemma-2-27b-it & 0.777 & \underline{0.714} & 0.607 & 0.682 & 0.648 & 0.623 & 0.653 & 0.578 & \textbf{0.766} \\
DeepSeek-V3 & 0.790 & \underline{0.578} & 0.552 & 0.557 & 0.557 & 0.572 & 0.574 & 0.460 &\textbf{ 0.607} \\
Qwen2.5-72B-Instruct & 0.796 & 0.500 & 0.509 & 0.573 & \underline{0.579} & 0.514 & 0.560 & 0.496 & \textbf{0.617} \\
\midrule
\multicolumn{1}{c}{\textbf{Average}} & 0.733 & 0.583 & 0.521 & \underline{0.597} & 0.580 & 0.533 & 0.578 & 0.521 & \textbf{0.628} \\

\bottomrule
\end{tabular}
    }
\end{table*}

\subsection{Uncertainty Percentage}\label{appendix:uncertainty_percentage}
\begin{figure*}
    \centering
    \includegraphics[width=\linewidth]{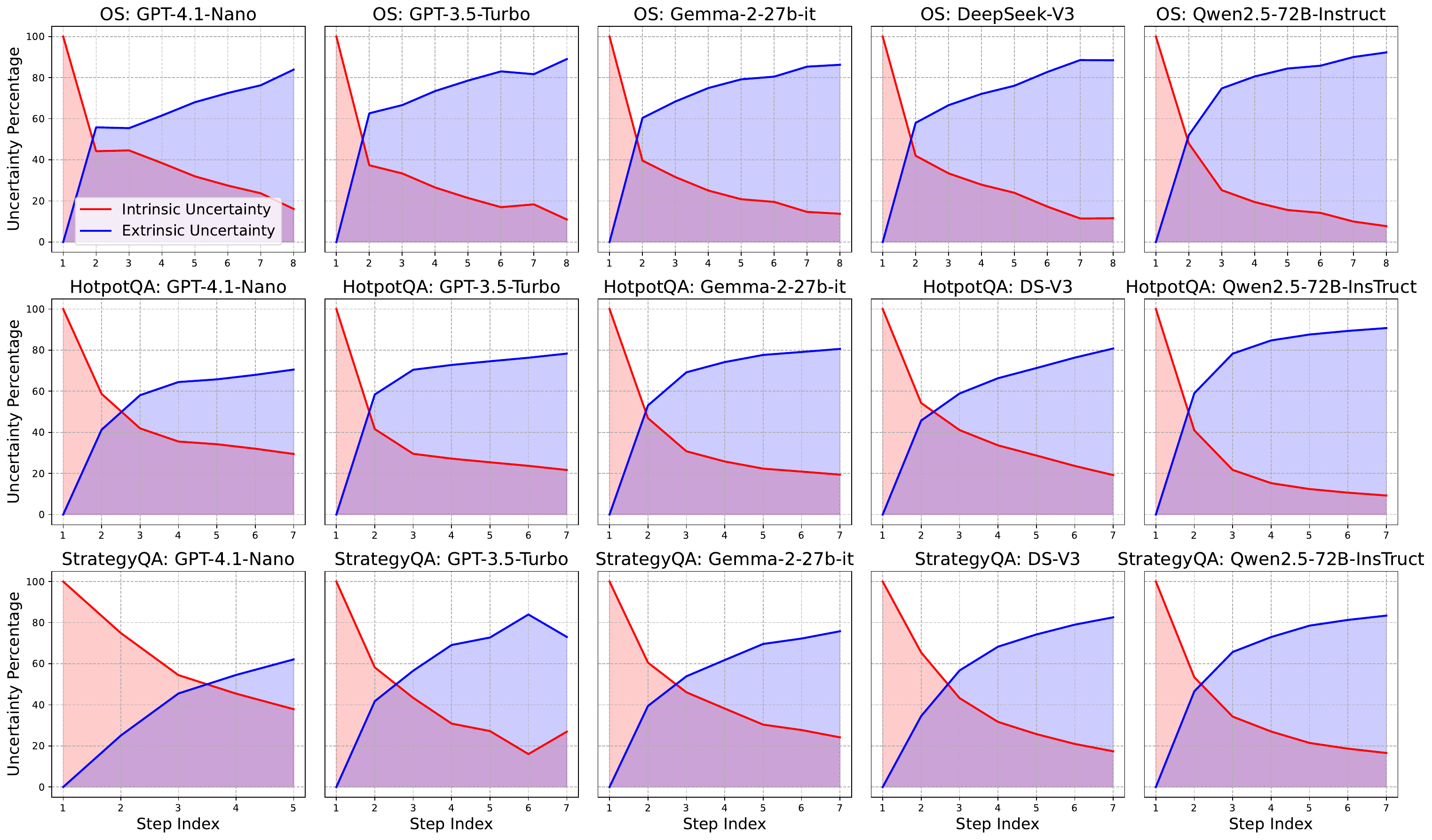}
    \caption{Detailed uncertainty percentage at each decision step.}
    \label{fig:detailed_unc_percentage}
\end{figure*}

In~\cref{fig:detailed_unc_percentage}, we provide the detailed uncertainty percentage at each model and benchmark.

%%%%%%%%%%%%%%%%%%%%%%%%%%%%%%%%%%%%%%%%%%%%%%%%%%%%%%%%%%%%

\end{document}